\documentclass{article} % For LaTeX2e
\usepackage{iclr2027_conference,times}

\usepackage{amsmath,amsfonts,bm}

\def\eqref#1{equation~\ref{#1}}
\def\1{\bm{1}}

\DeclareMathAlphabet{\mathsfit}{\encodingdefault}{\sfdefault}{m}{sl}
\SetMathAlphabet{\mathsfit}{bold}{\encodingdefault}{\sfdefault}{bx}{n}

\usepackage{hyperref}
\hypersetup{hidelinks, pdftitle={EvoUndo: Recoverability-Constrained Self-Evolution for LLM Agent Harnesses}, pdfauthor={Tanmay Sah, Dolly Sah, Harshul Jain, Tanya Sah}, pdfsubject={}, pdfkeywords={}}
\usepackage{url}
\usepackage{booktabs}
\usepackage{graphicx}

\title{EvoUndo: Recoverability-Constrained Self-Evolution for LLM Agent Harnesses}

\author{\normalfont\begin{tabular}[t]{@{}p{0.52\textwidth}p{0.41\textwidth}@{}}
\textbf{Tanmay Sah} & \textbf{Dolly Sah} \\
Independent Researcher & Independent Researcher \\
\texttt{tradertanmay@gmail.com} & \texttt{dolly17sah@gmail.com} \\[12pt]
\textbf{Harshul Jain} & \textbf{Tanya Sah} \\
Independent Researcher & Independent Researcher \\
\texttt{harshuljain1393@gmail.com} & \texttt{tanyasah20@gmail.com}
\end{tabular}}

\iclrfinalcopy

\begin{document}

\maketitle
\lhead{Preprint}

\begin{abstract}
LLM agents increasingly modify their own prompts, tools, middleware, resources, and execution harnesses at runtime. Such self-evolution can improve capability, but a successful mutation may leave persistent effects that cannot be safely reversed in states different from the one in which it was created. We introduce \textbf{EvoUndo}, a framework for representing, synthesizing, diagnosing, and independently verifying recoverability of model-generated self-modifications across counterfactual states. Across 600 unseen self-evolution tasks using single-sample zero-shot generation, we identify 197 capability-improving mutations that fail recoverability verification. Under the original recovery representation, conventional repair strategies recover 0/197 of these natural failures. Deterministic oracle analysis recovers 48/197 under the original recovery language $L_0$, while the extended recovery calculus increases empirical oracle recovery to 191/197. A protocol-locked $2\times2$ grounding-by-expressivity intervention then separates two bottlenecks: state-grounded diagnostic feedback increases successful recovery from 0/48 to 38/48 (79.2\%) when the original language is sufficient, while extending the recovery language enables recovery on 142/143 (99.3\%) failures in the oracle-defined $S_1$ stratum. On the primary gpt-oss-120b backbone, adding state-grounded diagnostics to the richer language reduces recovery to 133/143 (93.0\%); the negative interaction did not reproduce under the Qwen3.8-27B constrained-decoding replication configuration, which preserves the directions of the grounding and expressivity effects. These results indicate that reliable agent self-evolution requires co-designing verification, state grounding, witness semantics, and recovery-language expressivity rather than relying on iterative prompting alone.
\end{abstract}
 \section{Introduction}
\label{sec:intro}

LLM agents are increasingly becoming \emph{mutable systems}. Beyond selecting discrete actions, modern agents can modify prompts, install or remove tools, change middleware, alter configurations, allocate resources, register event listeners, and reconfigure their execution harnesses at runtime~\citep{yin2024godelagent,zhang2025dgm,lin2026ahe}. These capabilities create a path toward \emph{self-evolving agent harnesses}: systems that autonomously adapt their execution environment to novel tasks and failure modes.

Most self-improvement procedures naturally optimize the forward effect of a modification: given a harness state $H$ and a candidate mutation $m$, a candidate is favored whenever it improves a task objective, $J(m(H)) > J(H)$. However, forward improvement is insufficient for long-lived autonomous systems. A successful mutation may overwrite configuration values, reorder middleware chains, shadow existing tools, create temporary files, or leak background resources. If that mutation later becomes obsolete, causes a regression, or conflicts with subsequent updates, simply applying a static inverse operation may fail to restore the prior state. Crucially, the correct recovery procedure is often \emph{state-dependent}: restoring observational equivalence requires information that existed in the pre-mutation state.

This motivates a fundamental requirement for autonomous self-evolution: capability-improving mutations must also be \emph{robustly recoverable}. After applying a mutation to state $s$, the system must be able to utilize pre-mutation state evidence to restore a state observationally equivalent to $s$. Furthermore, this property must hold not merely in the nominal state used during construction, but across counterfactual states in which the mutation may operate.

We formalize and investigate this challenge: \emph{How should an autonomous agent search for capability-improving harness mutations when robust recoverability is an explicit invariant of the search space?}

We introduce \textbf{EvoUndo}, a framework for \emph{recoverability-constrained self-evolution}. EvoUndo provides a framework for representing, synthesizing, diagnosing, and independently verifying recoverability of model-generated self-modifications. Candidates are executed across counterfactual harness states and admitted only when the recovered states satisfy typed observational equivalence with their corresponding pre-states. Recovery failure is treated as an actionable defect to diagnose and repair before a mutation is persistently admitted.

A central insight of this work is that recoverability is \emph{relational}: whether a mutation is recoverable depends jointly on the state distribution, witness capture semantics, observational contract, and expressivity of the recovery language. On 120 deliberately corrupted but recoverable mutations, verifier-guided typed diagnosis repairs 114/120 (95.0\%) cases at budget four, compared with 4/120 (3.3\%) under independent regeneration. In sharp contrast, across 600 unseen self-evolution tasks using single-sample zero-shot generation, 197 capability-positive mutations fail recovery verification, and conventional verifier-guided repair recovers 0/197 under the original recovery representation.

A zero-generation solvability audit reveals that only 48/197 natural failures are recoverable by the deterministic oracle under the original language $L_0$, whereas an extended language $L_1$ raises empirical oracle recoverability to 191/197. A protocol-locked $2\times2$ grounding-by-expressivity factorial separates these failure modes: state-grounded diagnostic feedback raises recovery from 0/48 to 38/48 (79.2\%) when $L_0$ is sufficient, while $L_1$ unlocks recovery on 142/143 (99.3\%) failures in the oracle-defined $S_1$ stratum. On gpt-oss-120b, adding state-grounded diagnosis on top of $L_1$ reduces recovery to 133/143 (93.0\%), while this negative interaction did not reproduce under the Qwen3.8-27B constrained-decoding replication configuration.

\paragraph{Contributions.}
\begin{enumerate}
    \item We formalize \textbf{recoverability-constrained self-evolution} and introduce \textbf{EvoUndo}, coupling harness mutations with witness capture, counterfactual verification, typed diagnosis, and closed-loop recovery synthesis.
    \item We establish a natural failure benchmark from 600 self-evolution tasks, identifying 197 capability-positive recovery failures where standard iterative repair achieves 0/197 success under the base recovery representation.
    \item We decouple natural recovery failures into \textbf{grounding} and \textbf{expressivity} bottlenecks: state-grounded diagnostic feedback recovers 38/48 (79.2\%) failures in $S_0$, while an extended recovery calculus enables recovery on 142/143 (99.3\%) failures in $S_1$.
    \item We observe a non-monotonic interaction on the primary configuration between diagnostic granularity and recovery-language capacity: state-grounded feedback degrades recovery under $L_1$ on gpt-oss-120b, while this negative interaction did not reproduce under the Qwen3.8-27B constrained-decoding replication configuration.
\end{enumerate}
 \section{Problem Formulation}
\label{sec:problem}

\subsection{Self-Evolving Agent Harnesses}
We model an agent harness as $H=(s,\Pi)$, where $s\in\mathcal{S}$ is a concrete persistent harness state and $\mathcal{S}$ is the state space. The state includes configuration, prompt templates, routing, tool registries, middleware sequences, event-listener bindings, sandboxed files, and managed resources; $\Pi$ is the execution policy. A mutation has a state projection $m:\mathcal{S}\to\mathcal{S}$. At a nominal state $s_\star\in\mathcal{S}$, it is \emph{capability-improving} when $\Delta J(m;s_\star)=J(m(s_\star))-J(s_\star)>0$ for a forward objective $J$. We reserve $S_0,S_1$ for the empirical task strata below. Persistent modifications must carry verified recovery semantics before permanent admission.

\subsection{Witnessed Recovery and Example}
Because forward mutations may overwrite information, $m(s)$ is generally insufficient to reconstruct its predecessor. The pre-state capture program $w:\mathcal{S}\to\mathcal{W}_{\mathrm{pre}}$ records recovery-relevant prior values and existence flags. Forward execution additionally produces an allocation receipt:
\[
\lbrack\!\lbrack m\rbrack\!\rbrack(s)=(m(s),\mathcal{R}_m(s)),\qquad
\hat{s}=u(m(s),w(s),\mathcal{R}_m(s)).
\]
Here $\mathcal{R}_m(s)\in\mathcal{P}(\mathcal{K}_{\mathrm{res}}\times\mathcal{H})$ binds symbolic resource keys to dynamically acquired handles; $u$ consumes both the pre-state witness and this mutation-produced receipt. A pre-state witness cannot contain a future socket handle. Resource capture stores lightweight allocation descriptors, not serialized live sockets; symbolic recovery operations use the runtime registry to close newly allocated handles and remove bindings.

\emph{Example}: Consider a configuration mutation $m = \texttt{set\_config("timeout\_sec", 60)}$ applied to an initial state $s$ where $\texttt{config["timeout\_sec"]} = 30$. Because $m(s)$ contains only 60, reconstructing $s$ requires witness capture $w(s)$ recording pre-state value 30 and that the key existed. The recovery program $u$ executes $\texttt{restore\_config("timeout\_sec", witness\_value)}$, successfully restoring 30. If the key did not previously exist, recovery instead deletes the newly introduced binding. In contrast, consider inserting a rate limiter into a middleware chain: $m$ inserts a filter at index 0. Here, a simple scalar inverse is insufficient; recovery must capture pre-mutation sequence positions and execute ordered indexed restoration ($L_1$), illustrating why richer state surfaces demand structured recovery calculi.

\subsection{Typed Observational Equivalence}
Rather than demanding full byte-level equality over an entire operating environment, recovery is evaluated using state-specific \emph{observational equivalence} $\hat{s} \simeq_{\mathcal{C}_e} s$ governed by an effect contract $\mathcal{C}_e$. The contract declares typed state targets whose effects must be covered. Semantic representations are recursively canonicalized: dictionary keys are sorted, volatile identifiers/pointers are mapped to canonical placeholders, and file contents are verified via SHA-256 hashes (excluding ephemeral timestamps/inodes). Configuration mappings, tool parameter schemas, middleware ordering, listener callbacks, and resource descriptors are checked against their pre-mutation semantic definitions (details in Appendix~\ref{app:counterfactual}).

Let $\Sigma$ denote the modeled persistent state surfaces. Typed restoration rules apply only within $\mathcal{C}_e$; outside it, the verifier requires strict canonical equality. Writing $\hat{s}\simeq_{\mathcal{C}_e}s$ for this combined relation,
\begin{equation}
\begin{aligned}
s_{\mathrm{rec}}\equiv_{\mathcal{T}(\mathcal{C}_e)}s_{\mathrm{pre}}
\iff{}&\left(\forall\sigma\in\mathcal{C}_e,\;
 s_{\mathrm{rec}}[\sigma]\equiv_\sigma s_{\mathrm{pre}}[\sigma]\right)\\
&{}\land\left(\forall\sigma\in\Sigma\setminus\mathcal{C}_e,\;
 s_{\mathrm{rec}}[\sigma]=s_{\mathrm{pre}}[\sigma]\right).
\end{aligned}
\label{eq:typed_equivalence}
\end{equation}
Thus persistent recovery corruption outside $\mathcal{C}_e$ fails verification. An empty contract requires global canonical equality and cannot cover any nonempty forward effect $E(m,s)$. These checks concern final modeled persistent state: transient recovery-side effects that disappear before final-state verification are not independently constrained.

\subsection{Counterfactual Recoverability}
Recovery must generalize beyond the specific state on which a mutation was drafted. Let $\mathcal{L}$ denote the available recovery language and $Q$ a distribution over counterfactual harness states. We define counterfactual recoverability as:
\begin{equation}
\mathcal{R}_{\mathcal{L},Q,\mathcal{C}_e}(m) = \mathbf{1}\left[ \exists (w,u)\in\mathcal{L} : \Pr_{s\sim Q}\left[ u(m(s),w(s),\mathcal{R}_m(s)) \simeq_{\mathcal{C}_e} s \;\land\; E(m,s)\subseteq\mathcal{C}_e \right] \geq \tau_R \right],
\label{eq:recoverability}
\end{equation}
where $\tau_R \in (0, 1]$ is the required recovery threshold. Recoverability is therefore an inherently \emph{relational} property parameterized by the forward mutation, witness/recovery representations, observational contract, and state distribution $Q$.

\subsection{Recoverability-Constrained Objective}
The goal of recoverability-constrained self-evolution is to discover forward-improving mutations that satisfy robust counterfactual recovery:
\begin{equation}
\begin{aligned}
\max_{m,w,u,\mathcal{C}_e}\;&J(m(s_\star)) \\
\text{s.t.}\;&\Delta J(m;s_\star)>0,\quad (w,u)\in\mathcal{L},\quad \mathcal{C}_e\text{ valid},\\
&\Pr_{s\sim Q}\!\left[\hat{s}\simeq_{\mathcal{C}_e}s\;\land\;E(m,s)\subseteq\mathcal{C}_e\right]\geq\tau_R.
\end{aligned}
\label{eq:objective}
\end{equation}
Our recovery-repair experiments hold $m=m_0$ fixed and restrict search to $(w,u,\mathcal{C}_e)$. This intentional control isolates witness construction, recovery synthesis, and effect-contract reasoning while preventing apparent recovery gains from weakening forward capability. It also limits the study: joint optimization of forward mutation and recovery was not evaluated. Future work could explore capability-equivalent but more reversible forward mutations.
 \section{EvoUndo}
\label{sec:method}

\subsection{Candidate Representation and Effect Contracts}
An EvoUndo candidate is a 4-tuple $\xi = (m, w, u, \mathcal{C}_e)$, where $m$ is the forward mutation, $w$ is the witness program, $u$ is the recovery program, and $\mathcal{C}_e$ is the declared effect contract. In EvoUndo recovery repair, the forward mutation $m$ is strictly immutable ($m_{\mathrm{repaired}} \equiv m_0$): $m_0$ is locked at Turn 0 and executed verbatim. Only the witness capture program $w$, recovery program $u$, and declared effect contract $\mathcal{C}_e$ may be revised across repair turns. The contract is constrained to the runtime's typed effect-contract vocabulary; if omitted, the compiler infers it from the locked forward operations. Although mutable, $\mathcal{C}_e$ is not trusted as complete: the runtime independently computes $E(m,s)=\operatorname{SnapshotDiff}(s,m(s))\cup\operatorname{ExecutionTraceEffects}(m,s)$ and admission requires $E(m,s)\subseteq\mathcal{C}_e$. Omitting an observed effect therefore fails the contract audit, whereas over-declaration enlarges the equivalence scope that recovery must satisfy. The immutability of $m$ prevents evasion by weakening the forward mutation to a no-op, while independent effect tracking prevents evasion by shrinking the declared recovery scope.

\subsection{Counterfactual Round-Trip Verification}
For each counterfactual state $s \in \mathcal{Q} = \mathcal{Q}_{\text{dev}} \cup \mathcal{Q}_{\text{hid}}$, EvoUndo evaluates the round trip $w_{\mathrm{pre}}=w(s)$, $(s',\mathcal{R}_m(s))=\lbrack\!\lbrack m\rbrack\!\rbrack(s)$, and $\hat{s}=u(s',w_{\mathrm{pre}},\mathcal{R}_m(s))$, checking whether $\hat{s} \simeq_{\mathcal{C}_e} s$. Development states $Q_{\mathrm{dev}}$ generate state-level diagnostic feedback. In the original factorial, hidden state contents and per-state outcomes are withheld, although aggregate IID/OOD recovery LCB summaries are returned during repair. The fresh-holdout study instead uses a fully feedback-isolated hidden set for final admission.

\subsection{Diagnostic Feedback Granularity}
When a development round trip fails, the verifier computes the state residual $\Delta_{\text{res}} = \operatorname{Diff}_{\mathcal{C}_e}(s, \hat{s})$. Under \textbf{coarse typed diagnosis} ($D_0$), the model receives the affected subsystem and defect classification without exact residual locations. Under \textbf{state-grounded diagnosis} ($D_1$), the feedback supplies a composite bundle of canonical state addresses, forward effect ordering, observed residuals, residual counts, and verifier observations (prompts in Appendix~\ref{app:factorial}). Causal interpretation applies to this bundle as a whole; addresses were not isolated from the other diagnostic fields.

\subsection{Recovery Languages}
Synthesis is bounded by a formal recovery language $\mathcal{L}$. Table~\ref{tab:dsl_grammar} gives the implemented primitives and operational semantics. Witness capture runs before $m$, recording typed pre-state values and existence flags in $w(s)$; recovery also consumes mutation-produced resource receipts. A primitive restores the previous binding if it existed, or deletes/unregisters a newly introduced binding if it did not. $L_0$ covers configuration, prompt, tool-registry, and routing state. $L_1$ extends it with middleware sequences, event listeners, sandboxed files, managed sockets, and ordered composition of recovery operations across surfaces. For example, a mutation that prepends middleware and registers a listener is recovered by unregistering/restoring the listener before restoring the prior middleware sequence in strict Last-In, First-Out (LIFO) order relative to the forward operations. The runtime rejects any $L_1$-only primitive emitted under $L_0$ and consumes that synthesis attempt. These are implemented operational semantics, not a completeness characterization of recovery programs.

\begin{table}[ht]
\centering
\footnotesize
\caption{Recovery language calculi ($L_0$ and $L_1$) over modeled persistent state targets.}
\label{tab:dsl_grammar}
\vspace{0.15cm}
\begin{tabular}{llp{0.49\textwidth}p{0.20\textwidth}}
\toprule
\textbf{Language} & \textbf{Stage} & \textbf{Primitive(s)} & \textbf{Target / state role} \\
\midrule
$L_0$ & Capture & \raggedright \texttt{capture\_config}, \texttt{capture\_prompt}, \texttt{capture\_tool}, \texttt{capture\_routing} & Typed pre-state witness \\
$L_0$ & Forward & \raggedright \texttt{set\_config}, \texttt{set\_prompt}, \texttt{register\_tool}, \texttt{set\_routing} & Locked forward effect \\
$L_0$ & Recovery & \raggedright \texttt{restore\_config}, \texttt{delete\_config}, \texttt{restore\_prompt}, \texttt{restore\_tool}, \texttt{unregister\_tool}, \texttt{restore\_routing} & Restore or remove prior binding \\
\midrule
$L_1$ & Capture & \raggedright \texttt{capture\_middleware}, \texttt{capture\_listener}, \texttt{capture\_file}, \texttt{capture\_socket} & Structured pre-state witness \\
$L_1$ & Forward & \raggedright \texttt{add\_middleware}, \texttt{add\_listener}, \texttt{write\_file}, \texttt{allocate\_socket} & Locked structural effect \\
$L_1$ & Recovery & \raggedright \texttt{restore\_middleware}, \texttt{restore\_listener}, \texttt{unregister\_listener}, \texttt{restore\_file}, \texttt{delete\_file}, \texttt{release\_socket} & Structured restoration or removal \\
\bottomrule
\end{tabular}
\end{table}

\subsection{Closed-Loop Synthesis and Admission}
If a capability-positive mutation fails recovery verification, EvoUndo initiates bounded synthesis over $(w, u, \mathcal{C}_e)$ up to budget $B \in \{1, 2, 3, 4\}$ with early stopping while keeping $m_0$ fixed. Proposals passing $\mathcal{Q}_{\text{dev}}$ are evaluated on $\mathcal{Q}_{\text{hid}}$. The executed primary protocol uses separate IID and OOD hidden splits ($n=20$ each). Each independently requires the Wald normal-approximation lower endpoint $\max(0,\hat p-1.96\sqrt{\hat p(1-\hat p)/n})\geq\tau_R=0.85$, where $\hat p=k/n$. Thus 20/20 and 19/20 pass, while 18/20 fails: at least 19/20 successes are required on each split. Post-hoc pooled $N=40$ two-sided 95\% Wilson rescoring at threshold 0.85 gives identical decisions for all recorded candidate evaluations; it was not the executed protocol (Appendix~\ref{app:admission_audit}). EvoUndo fails closed: syntax errors, unhandled exceptions, or out-of-language primitives result in immediate rejection, ensuring every admitted state transition satisfies the configured empirical counterfactual recoverability criterion over the modeled harness state.
 \section{Experimental Setup}
\label{sec:setup}

\subsection{Harness State and Task Benchmark}
The natural benchmark comprises six 100-task architectural families ($N=600$): \textbf{Configuration}, \textbf{Tools}, \textbf{Middleware}, \textbf{Listeners}, \textbf{Resources}, and \textbf{Multi-Surface}. Multi-Surface is a composite benchmark family, not an atomic harness surface. These evaluation groupings do not enumerate persistent state structures one-to-one: prompt and routing state are also modeled; Resources covers file- and resource-backed mutations; and Multi-Surface tasks combine mutations touching at least two persistent state targets. Forward capability $J_t(m(s_\star))$ evaluates postcondition satisfaction for each task $t$ independently of recovery syntax. Config has zero recovery-defective tasks in the frozen primary cohort (Table~\ref{tab:family_distribution}).

\subsection{Model, Inference, and Verification Protocol}
Factorial repair and synthesis experiments use the open-weights \texttt{gpt-oss-120b} model under native MXFP4 precision at temperature 0.2 with medium reasoning effort on eight NVIDIA H200 GPUs. In the primary factorial study on the frozen 197-task cohort, each candidate is evaluated across $|\mathcal{Q}_{\text{dev}}|=10$ development states and $|\mathcal{Q}_{\text{hid}}|=40$ hidden states (20 IID, 20 OOD), gated independently by the split-Wald rule in Section~\ref{sec:method}. IID states perturb pre-state boundary values, prior-existence states (absent vs.\ conflicting tool schemas), and combinatorial surface assignments. OOD states introduce true structural shifts: unfamiliar nested keys, conflicting socket allocations, prepended high-priority middleware, multi-level file paths, and reverse routing defaults (detailed taxonomy in Appendix~\ref{app:counterfactual}). Generalization is further verified on an untouched 300-task fresh holdout with 100-state hidden evaluation (Section~\ref{sec:fresh}). In a separate controlled audit of 300 intentionally incomplete effect contracts across all six harness families, independent dynamic effect tracking detected every injected undeclared semantic effect (300/300), with no false positives on 300 matched complete contracts; Appendix~\ref{app:counterfactual} provides details.

\subsection{Controlled Benchmark and Natural Failure Bank}
We construct a \textbf{controlled benchmark} of 120 capability-positive mutations with injected witness/recovery defects and a \textbf{natural failure bank} by single-sample zero-shot generation across 600 tasks. Of these, 281 are admissible, 122 are capability-inadequate, and \textbf{197} are capability-positive but fail recovery verification, forming our frozen primary cohort. Turn-0 generation used one candidate per task, zero demonstrations, and no rerolls; complete generation parameters are reported in Appendix~\ref{app:natural}.

\subsection{Solvability Audit and Mechanistic Strata}
To distinguish synthesis search failures from empirical representational limits under the implemented recovery languages, we perform a zero-generation constructive oracle audit. Under base language $L_0$, only 48/197 mutations are oracle-recoverable. Under extended language $L_1$, 191/197 are oracle-recoverable. We define empirical oracle recoverability $\mathcal{R}_{\mathcal{L}, Q}^{\mathrm{oracle}}(m) \in \{0, 1\}$ as successful recovery synthesis and hidden-state admission by the deterministic constructive oracle under language $\mathcal{L}$. This partitions the failure bank into two disjoint mechanistic strata:
\begin{align}
S_0 &= \{m : \mathcal{R}_{L_0, Q}^{\mathrm{oracle}}(m) = 1\}, \quad |S_0| = 48, \\
S_1 &= \{m : \mathcal{R}_{L_1, Q}^{\mathrm{oracle}}(m) = 1 \land \mathcal{R}_{L_0, Q}^{\mathrm{oracle}}(m) = 0\}, \quad |S_1| = 143.
\end{align}
Six tasks remain unrecovered because of implemented oracle heuristics rather than demonstrated theoretical impossibility and are excluded from stratum contrasts (Appendix~\ref{app:oracle_admission}). The oracle uses privileged pre-state and evaluator information only for retrospective stratification; no model receives its programs or hidden outcomes. It constructs its own tight effect contract from forward-operation surfaces and observed mutation effects, independently of candidate declarations. Stratum membership was unchanged across 10 independent resamplings from the implemented counterfactual generator (Appendix~\ref{app:stratum_stability}).

\subsection{Grounding--Expressivity Factorial and Contrasts}
We evaluate a $2\times2$ factorial crossing diagnostic granularity with recovery language: $\{D_0, D_1\} \times \{L_0, L_1\}$. Before execution, we protocol-locked three confirmatory paired contrasts:
\begin{itemize}
    \item $C_1$: $D_1L_0$ vs.\ $D_0L_0$ on $S_0$ (tests state-grounded diagnostic feedback when $L_0$ is sufficient).
    \item $C_2$: $D_0L_1$ vs.\ $D_0L_0$ on $S_1$ (tests recovery-language expressivity on the $S_1$ stratum).
    \item $C_3$: $D_1L_1$ vs.\ $D_0L_1$ on $S_1$ (tests diagnostic grounding on top of an expressive language).
\end{itemize}
We report Rescue@$B$ (the fraction of recovery-defective tasks successfully admitted within at most $B$ sequential synthesis turns, stopping early on success; $B=4$ in the primary factorial), paired risk differences $\Delta$ with 95\% bootstrap CIs, and exact two-sided McNemar tests with Holm--Bonferroni correction. Extended related work is detailed in Appendix~\ref{app:related}.
 \section{Controlled and Natural Recovery Results}
\label{sec:results}

\subsection{Controlled Recoverability: Repair Succeeds When Representation Aligns}
On the 120-task controlled benchmark, independent regeneration recovers only 4/120 (3.3\%) defects at budget $B=4$. In contrast, closed-loop verifier feedback achieves high repair rates: generic feedback recovers 104/120 (86.7\%), raw verifier traces recover 101/120 (84.2\%), typed diagnosis recovers 114/120 (95.0\%), and prescriptive hints reach 117/120 (97.5\%). Thus, when defects fall within the language's representational envelope, closed-loop repair is highly effective.

\subsection{Natural Failures Break Conventional Repair}
On the frozen cohort of 197 natural self-evolution failures, this success collapses entirely. Under the base representation ($L_0$), all four verifier-guided repair modes achieve \textbf{0/197 (0.0\%)} recovery at $B=4$, while independent regeneration recovers 6/197 (3.0\%). Baseline taxonomy classification identified 180/197 natural failures outside the predefined controlled defect taxonomy. This set is distinct from the 180/197 tasks rescued by $D_0L_1$: the overlap is 164 tasks, with 16 taxonomy-only and 16 rescue-only tasks. Equal cardinalities do not identify the same tasks.

\paragraph{End-to-end 600-task accounting.}
Of all 600 unseen tasks, 478/600 (79.67\%) produced capability-positive mutations: 281 were already admissible and 197 were recovery-defective (197/478, 41.2\% of capability-positive mutations). Under $D_0L_1$ at $B=4$, EvoUndo rescues 180/197 (91.37\%) defects, yielding \textbf{461/478 (96.44\%)} admissible capability-positive mutations and \textbf{461/600 (76.83\%)} total benchmark yield. Relative to the initially admissible 281/600 (46.83\%), this is a $+30.00$ pp absolute increase. Because recovery repair executes the frozen $m_0$ verbatim and revises only $(w,u,\mathcal{C}_e)$, all 180 rescued candidates retain their original positive forward capability; zero forward-capability regressions were introduced by recovery repair. This fixed-$m_0$ result does not establish the capability--recoverability trade-off under joint optimization with $m$.

\begin{table}[t]
\centering
\small
\caption{Final protocol-locked Grounding--Expressivity Factorial results on the frozen natural failure cohort ($N=197$) at budget $B=4$. $S_0$ ($N=48$) denotes failures recovered by the deterministic oracle under $L_0$; $S_1$ ($N=143$) denotes failures recovered under $L_1$ but not by the $L_0$ oracle.}
\label{tab:factorial}
\vspace{0.15cm}
\resizebox{\textwidth}{!}{
\begin{tabular}{lccc}
\toprule
\textbf{Factorial Condition} & \textbf{Full Cohort} ($N=197$) & \textbf{Stratum $S_0$} ($N=48$) & \textbf{Stratum $S_1$} ($N=143$) \\
\midrule
$D_0L_0$ (Coarse Diagnosis, Base Language) & 0/197 (0.0\%) & 0/48 (0.0\%) & 0/143 (0.0\%) \\
$D_1L_0$ (Grounded Bundle, Base Language) & 38/197 (19.3\%) & 38/48 (79.2\%) & 0/143 (0.0\%) \\
$D_0L_1$ (Coarse Diagnosis, Rich Language) & \textbf{180/197 (91.4\%)} & 38/48 (79.2\%) & \textbf{142/143 (99.3\%)} \\
$D_1L_1$ (Grounded Bundle, Rich Language) & 167/197 (84.8\%) & 34/48 (70.8\%) & 133/143 (93.0\%) \\
\bottomrule
\end{tabular}
}
\end{table}

\section{Grounding--Expressivity Factorial Results}
\label{sec:factorial}

Table~\ref{tab:factorial} reports the results of the $2\times2$ factorial experiment.

\subsection[C1: Diagnostic Grounding Unlocks Recoveries on Stratum S0]{C1: Diagnostic Grounding Unlocks Recoveries on Stratum $S_0$}
On stratum $S_0$, coarse diagnosis ($D_0L_0$) yields 0/48 recoveries. Supplying the state-grounded diagnostic bundle ($D_1L_0$) while keeping the base language $L_0$ fixed increases successful recovery to \textbf{38/48 (79.2\%)}. The paired risk difference is $\Delta = +79.17$ pp (95\% bootstrap CI $[+66.67, +89.58]$), with exact McNemar $p_{\text{raw}} = 2^{-37} = 7.28\times 10^{-12}$ ($p_{\text{Holm}} = 1.46\times 10^{-11}$). For tasks in $S_0$, resolving the \emph{grounding bottleneck} enables the model to locate and restore the affected state surfaces.

\subsection[C2: Language Expressivity Unlocks Recoveries on Stratum S1]{C2: Language Expressivity Unlocks Recoveries on Stratum $S_1$}
On stratum $S_1$, which contains tasks recovered by the deterministic $L_1$ oracle but not by the $L_0$ oracle, $D_0L_0$ yields 0/143 recoveries, whereas $D_0L_1$ recovers \textbf{142/143 (99.3\%)}. The paired risk difference is $\Delta = +99.30$ pp (95\% CI $[+97.90, +100.00]$, $p_{\text{Holm}} = 1.08\times 10^{-42}$). This demonstrates that providing adequate recovery primitives overcomes the \emph{expressivity bottleneck}, allowing the model to synthesize valid recovery programs for nearly all tasks in $S_1$.

\subsection{C3: State-Grounded Diagnostics Interact Non-Monotonically with Language Capacity}
Crucially, adding state-grounded feedback on top of the expressive language $L_1$ does not improve performance. On $S_1$, recovery drops from 142/143 (99.3\%) under $D_0L_1$ to \textbf{133/143 (93.0\%)} under $D_1L_1$. The paired risk difference is $\Delta = -6.29$ pp (95\% CI $[-11.19, -2.10]$, exact McNemar $p_{\text{Holm}} = 0.0117$). Trace analysis of the 11 discordant tasks suggests changes in semantic decomposition and operation ordering rather than systematically longer recoveries; the detailed taxonomy and generation metrics appear in Appendix~\ref{app:discordant}. This analysis is descriptive rather than causal.

\subsection{Synthesis Efficiency and Interaction}
Synthesis efficiency is detailed in Appendix~\ref{app:synthesis_efficiency}.

The full-cohort interaction on gpt-oss-120b is $I=(D_1L_1-D_0L_1)-(D_1L_0-D_0L_0)=-25.89$ pp (95\% CI $[-34.01,-18.27]$): the diagnostic bundle benefits restrictive $L_0$ but is not additive under $L_1$.

\paragraph{Cross-model replication.}
A frozen 96-task Qwen3.8-27B replication with constrained structured decoding preserves the positive grounding and expressivity effects (Table~\ref{tab:qwen_main}; full cells in Appendix~\ref{app:qwen}).

\begin{table}[t]
\centering
\small
\caption{Cross-model mechanistic contrasts on the frozen 96-task Qwen3.8-27B replication.}
\label{tab:qwen_main}
\vspace{0.1cm}
\begin{tabular}{llcc}
\toprule
\textbf{Contrast} & \textbf{Mechanism} & \textbf{$\Delta$ (pp)} & \textbf{Holm-adjusted $p$} \\
\midrule
$C_1$: $D_1L_0-D_0L_0$ on $S_0$ & Diagnostic grounding & $+47.92$ & $<0.001$ \\
$C_2$: $D_0L_1-D_0L_0$ on $S_1$ & Language expressivity & $+37.50$ & $<0.001$ \\
$C_3$: $D_1L_1-D_0L_1$ on $S_1$ & Added grounded diagnostics & $+6.25$ & $0.375$ \\
\bottomrule
\end{tabular}
\end{table}

The negative $C_3$ interaction observed on gpt-oss-120b did not reproduce under the Qwen3.8-27B constrained-decoding replication configuration ($+6.25$ pp, 95\% CI $[-2.08,+16.67]$, $p_{\mathrm{adj}}=0.375$). Model, decoding protocol, and sample size differ, so the experiment does not isolate the cause.

\section{Fresh-Holdout Validation and Robustness}
\label{sec:fresh}
We evaluated frozen $L_1$ on \textbf{300 fresh tasks} (50 per family) with 100 hidden states. Increasing diverse development states from $Q_2$ to $Q_{20}$ reduced observed false positives from 16.8\% to 0.0\%, but lowered $\mathrm{Rescue@4}_{100}$ from 49.7\% to 35.0\% ($\Delta=-14.67$ pp, $p=5.76\times10^{-6}$) and increased the conservatism gap from 6.6\% to 24.1\%.

The strict development gate requires perfect round trips (10/10 in the primary factorial; 20/20 under $Q_{20}$), potentially rejecting highly reliable candidates. Under $Q_{20}$, Listeners had 0/50 development passes but 32/50 hidden-criterion passes. This precision--coverage trade-off remains unresolved; zero observed false positives does not guarantee zero population risk. Appendix~\ref{app:fresh} gives the rejection-probability illustration and family breakdown.

Additional sweeps over development-set composition, hidden sample size, recovery threshold, and synthesis budget preserve the qualitative precision--coverage trade-off; full results appear in Appendix~\ref{app:fresh}.

Fresh-holdout inference cost ranges from 2.39--3.09 calls and 3.37k--4.30k generated tokens per task; detailed latency and per-success costs appear in Appendix~\ref{app:fresh}. The final strict-$L_0$ $D_1L_0$ run averages 3.51 calls/task.
 \section{Discussion and Limitations}
\label{sec:discussion}
Under the evaluated medium-reasoning configuration and budget $B=4$, conventional verifier-guided repair did not overcome the observed representation-bounded failures under $L_0$ (0/197). No low-, medium-, or high-effort comparison was performed, so the study does not establish the effect of scaling reasoning effort. Moreover, diagnostic feedback interacts non-monotonically with language expressivity on the primary model. These results argue against treating maximal diagnostic specificity as universally preferable; diagnostic granularity should be treated as a configuration-dependent synthesis parameter rather than assumed to be uniformly beneficial. An adaptive diagnostic strategy---escalating from coarse failure indicators to granular state traces only when coarse repair fails---is a plausible mitigation, but no diagnostic curriculum or address-use constraint was evaluated here.

\paragraph{Snapshot and transactional recovery.}
Effect-scoped snapshots are the stronger baseline in our evaluated serializable in-memory selective-undo regime: they recover 300/300 tasks versus 243/300 for EvoUndo under different-surface changes and 159/300 versus 131/300 under same-surface changes (Appendix~\ref{app:snapshots}). There are no EvoUndo-only wins. EvoUndo targets settings where direct exact effect-scoped restoration is unavailable or insufficient; superiority there is not established. Reported rollback latency measures execution of an already-synthesized recovery and excludes LLM synthesis. No WAL or ARIES runtime was implemented or evaluated.

\emph{Limitations and Data Governance}: (1) $L_0/L_1$ are specialized to modeled state types and have no completeness guarantee. (2) Repair fixes $m=m_0$; joint forward/recovery optimization is untested. (3) Distributed and external state, including third-party APIs and unmanaged processes, is unmodeled. (4) Irreversible effects require compensation. (5) Only gpt-oss-120b and Qwen3.8-27B were evaluated. (6) Production witnesses require access controls, encryption, finite retention, least-privilege schemas, and redaction.

\section{Conclusion}
\label{sec:conclusion}
We introduced EvoUndo, demonstrating that reliable self-evolution requires more than forward capability improvement: verification, state grounding, witness semantics, and recovery-language expressivity must be designed together. When recovery representations align with the underlying state structure, LLMs can reliably synthesize state-dependent recovery programs across counterfactual states, providing a principled foundation for auditable, recoverable self-evolving autonomous agents.

\subsubsection*{AI Use Statement}
Generative AI tools were used for coding assistance and draft editing.

\subsubsection*{Reproducibility Statement}
All code, task definitions, counterfactual generators, protocol locks, and evaluation traces will be released openly. Appendices~\ref{app:related}--\ref{app:integrity} provide complete architectural schemas, extended related work, language specifications, prompts, baseline comparisons, and statistical derivations.

\subsubsection*{Ethics Statement}
This work develops safety constraints for autonomous agents. Experiments run in isolated sandboxes without human subjects or sensitive production data. Deployment in live infrastructure requires additional authorization and audit controls.

\clearpage
\bibliography{iclr2027_conference}
\bibliographystyle{iclr2027_conference}

\newpage
 \appendix

\section{Extended Related Work}
\label{app:related}

\paragraph{Self-Improving and Self-Evolving Agents.}
The concept of systems that iteratively modify their own internal behavior dates back to Schmidhuber's G\"{o}del machine formulation of self-referential improvement~\citep{schmidhuber2003goedel}. Recent advances in LLMs have made autonomous self-modification empirically practical. G\"{o}del Agent enables an LLM agent to recursively modify its own logic and optimization procedure~\citep{yin2024godelagent}, while the Darwin G\"{o}del Machine maintains an evolving population of self-modified coding agents and retains modifications according to downstream task performance~\citep{zhang2025dgm}. More recent work extends self-improvement from model weights or monolithic agent code to the surrounding executable harness. Self-Harness iteratively mines weaknesses, proposes harness modifications, and validates them using held-out performance~\citep{zhang2026selfharness}, while Agentic Harness Engineering uses explicit editable components, trajectory-derived evidence, and outcome validation to evolve coding-agent harnesses~\citep{lin2026ahe}. Recent work further treats the executable harness itself as an optimization target. Retrospective Harness Optimization (RHO) improves an agent harness from prior trajectories using self-validation and self-preference, without requiring an external labeled validation set~\citep{pan2026rho}. Co-Harness jointly alternates harness optimization with model-parameter optimization, allowing the runtime scaffolding and the underlying model to co-evolve~\citep{chen2026coharness}. Harness-R1 instead learns a dedicated harness-engineering policy that converts batches of agent failures into validated executable runtime patches optimized for downstream task success~\citep{shao2026harnessr1}. Related approaches optimize specific parts of the agent system: Agent symbolic learning treats prompts, tools, and pipeline structure as learnable symbolic parameters and updates them using language-based analogues of optimization~\citep{zhou2024symbolic}; EvoTool evolves modular tool-use policies through trajectory-grounded failure attribution, targeted mutation, and population selection~\citep{yang2026evotool}; and Tool-R0 uses self-play to co-evolve task generation and tool-use capability without an initial supervised task dataset~\citep{acikgoz2026toolr0}. Across both holistic harness optimization and component-level methods, existing work optimizes forward utility or policy capability; EvoUndo addresses the complementary question of whether persistent model-generated edits retain independently verified recovery semantics.

\paragraph{Reversible Agent Execution and Runtime Effects.}
Recent agent runtimes have begun treating reversibility as a first-class systems property. GoEX develops a runtime for autonomous LLM actions centered on post-facto validation, providing undo mechanisms and damage confinement so that executed actions can either be reverted or have their possible impact bounded~\citep{patil2024goex}. Shepherd represents agent--environment interactions as typed effects and supports scoped fork, replay, discard, and rollback of reversible execution state~\citep{yu2026shepherd}. Related work on spatiotemporal composability formalizes dynamic software components whose effects can be reverted when components are removed, providing a foundation for dynamically loadable agent harnesses~\citep{shi2026spatiotemporal}. These systems provide runtime mechanisms for reversible or compensable effects. EvoUndo studies a distinct challenge that arises when the modification itself is synthesized by an LLM: the required recovery procedure may be unknown, state-dependent, or inexpressible in the available recovery language. We therefore treat witness capture and recovery synthesis as objects of verification rather than assuming that correct inverse semantics are supplied by the component author or runtime.

\paragraph{Reversible Programming and Bidirectional State Transformation.}
Reversible computation and bidirectional programming study transformations for which information required to reconstruct prior states is preserved or explicitly represented. Reversible-effect systems extend these ideas to stateful computations and mutable effects~\citep{heunen2018reversible}. Classical bidirectional transformations, particularly lenses, formalize paired transformations that propagate updates between related state representations while satisfying consistency laws~\citep{foster2007lenses}. EvoUndo shares the need to retain sufficient prior-state information for reverse state transformation, but differs because its forward edits and recovery programs are model-generated, and recovery correctness is evaluated empirically across counterfactual harness states. Consequently, a non-injective forward mutation need not be rejected when a state-dependent witness preserves the information required for recovery.

\paragraph{Verification of Agent Modifications.}
Self-improving agents commonly validate candidate modifications through downstream benchmarks, regression tests, or deterministic evaluation, including population-based self-modification~\citep{zhang2025dgm} and harness optimization~\citep{zhang2026selfharness,lin2026ahe}. Such validation can determine whether a modification improves capability without establishing whether its persistent effects can be removed. EvoUndo separates these properties: forward capability is evaluated through the task objective, while recovery is independently evaluated through counterfactual round trips; development states provide diagnostics, while hidden-state contents and per-state outcomes are withheld. In the primary factorial, aggregate hidden IID/OOD LCB summaries are nevertheless returned during repair. This distinction is central: a mutation may improve task performance and pass ordinary regression tests while still destroying information, leaking state, or requiring recovery operations unavailable in the current representation.

\paragraph{Formal Verification of Self-Evolving Agents.}
SEVerA studies self-evolving agent synthesis under formal correctness constraints, using formally guarded generative-model calls and program verification to ensure that synthesized agents satisfy specified contracts independently of learned model parameters~\citep{banerjee2026severa}. This provides stronger deductive guarantees than EvoUndo's empirical counterfactual verification. The two approaches address complementary properties: SEVerA verifies that generated agent programs satisfy formal behavioral constraints, whereas EvoUndo asks whether persistent, model-generated harness mutations can be returned to their prior observational state using state-dependent witness and recovery semantics.

\paragraph{Runtime Safety Enforcement.}
AARM treats agent action execution as a security boundary and proposes a runtime control layer that intercepts actions before execution, evaluates them against authorization and intent policies, and records execution evidence~\citep{errico2026aarm}. AgentSpec similarly introduces a lightweight domain-specific language for specifying and enforcing runtime safety constraints on LLM agents through explicit triggers, predicates, and enforcement actions~\citep{wang2025agentspec}. VIGIL similarly converts behavioral specifications into executable runtime policies over agent--tool traces, including temporal, argument, and value-flow constraints~\citep{li2026vigil}. Such systems determine whether an action sequence satisfies a behavioral policy; EvoUndo targets a different temporal property: whether a persistent harness self-modification can later be removed while restoring the relevant prior harness state. Runtime policy enforcement and recoverability-constrained evolution can therefore be composed. Persistent self-evolution also creates security risks beyond per-action authorization: Zombie Agents demonstrates that attacker-controlled content can be incorporated into an agent's evolving long-term memory and persist across sessions, later inducing unauthorized behavior~\citep{yang2026zombie}. EvoUndo studies whether persistent harness modifications carry recovery semantics that permit their effects to be independently verified and removed.

\paragraph{Proof-based verification and runtime enforcement.}
Astrogator verifies LLM-generated Ansible programs against a formal query using symbolic interpretation, providing deductive guarantees within its specified automation calculus~\citep{councilman2025astrogator}. EvoUndo instead provides empirical typed round-trip evidence over heterogeneous modeled harness states; this covers recovery properties without claiming proof-level completeness. AgentSpec enforces per-action runtime policies through triggers, predicates, and enforcement actions~\citep{wang2025agentspec}; such governance is orthogonal to whether an admitted persistent mutation can later restore prior state, so the two mechanisms can be composed.

\section{EvoUndo Implementation Details}
\label{app:implementation}

\subsection{Harness State Representation}
We represent the agent harness as a typed persistent state containing configuration, prompt templates, routing state, tool registries, middleware sequences, event-listener bindings, sandboxed files, and managed resources. Each state target is associated with a canonical addressing scheme and a type-specific observational equivalence policy. The global frame condition in Equation~\ref{eq:typed_equivalence} additionally checks all modeled persistent state outside the effect contract.

\subsection{Candidate Representation}
Each EvoUndo candidate is represented as $\xi=(m,w,u,\mathcal{C}_e)$, where $m$ is the frozen forward mutation, $w$ is the witness-capture program, $u$ is the recovery program, and $\mathcal{C}_e$ is the effect contract. In EvoUndo recovery repair, the forward mutation $m$ is strictly immutable ($m_{\mathrm{repaired}} \equiv m_0$). Only the witness capture program $w$, recovery program $u$, and declared effect contract $\mathcal{C}_e$ may be revised across repair turns. The contract is constrained to the runtime's typed effect-contract vocabulary; when omitted, it is inferred from the locked forward operations. The runtime independently computes $E(m,s)=\operatorname{SnapshotDiff}(s,m(s))\cup\operatorname{ExecutionTraceEffects}(m,s)$ and requires $E(m,s)\subseteq\mathcal{C}_e$: an omitted observed effect causes rejection, while over-declaration expands the state scope that must satisfy recovery equivalence.

\paragraph{Witness and resource execution.}
The capture program records the prior value and a boolean \texttt{\{key\}\_existed}. A recovery primitive dynamically restores the old binding when this flag is true and deletes/unregisters the mutation-introduced binding when it is false; this is not an unconditional inverse. For managed resources, $w(s)$ stores lightweight prior-allocation descriptors, while forward execution appends allocation receipts to \texttt{created\_resources}. Recovery consumes both components and resolves symbolic resource keys through the runtime registry before explicit handle closure and unbinding. No future handle is attributed to the pre-state witness.

\subsection{Canonical State Addresses}
Verifier diagnostics use canonical semantic state locations rather than raw implementation-specific object identities. Representative addresses include \texttt{tools["tool\_id"]}, \texttt{config["key"]}, \texttt{middleware["middleware\_id"]}, \texttt{files["path"]}, and \texttt{resources["resource\_id"]}. Canonicalization is applied uniformly across verifier residual extraction and diagnostic rendering. Diagnostic addresses exposed to the recovery model are derived only from verifier-visible development executions.

\section{Recovery Languages}
\label{app:dsl}

\subsection[Base Recovery Language L0]{Base Recovery Language $L_0$}
The original recovery language supports local operations for configuration, prompt, tool-registry, and routing state. Its implemented capture, forward, and recovery primitives are enumerated in Table~\ref{tab:dsl_grammar}. The $L_0$ runtime explicitly rejects operations requiring middleware-sequence, listener, file-prestate, socket-descriptor, or ordered multi-surface semantics.

\subsection[Extended Recovery Language L1]{Extended Recovery Language $L_1$}
$L_1$ strictly extends $L_0$ with five implemented capability classes identified by the natural-failure solvability audit:
\begin{enumerate}
    \item \textbf{Indexed sequence capture and restoration}: preserves both sequence membership and ordering information.
    \item \textbf{Listener capture and restoration}: records the prior event binding and restores it with \texttt{OpSpec(op\_type="restore\_listener", target="<event\_name>", witness\_key="<w\_key>")} or removes a newly registered listener with \texttt{OpSpec(op\_type="unregister\_listener", target="<event\_name>")}.
    \item \textbf{File pre-state capture and restoration}: captures whether a file existed and, when necessary, its previous contents or state.
    \item \textbf{Resource-descriptor capture and restoration}: records prior binding existence and lightweight allocation descriptors; recovery also consumes the dynamic receipt produced by forward execution. Live sockets are not serialized into the pre-state witness.
    \item \textbf{Ordered multi-surface recovery}: permits recovery of interacting state mutations in an explicitly specified order.
\end{enumerate}
The runtime enforces the requested language level. An $L_1$-only primitive emitted in an $L_0$ condition is classified as an invalid model output and consumes the corresponding synthesis attempt. These are implemented operational semantics and do not establish formal language completeness.

\section{Counterfactual Verification and Audit Baselines}
\label{app:counterfactual}

\subsection{Round-Trip Evaluation and Counterfactual Generation Taxonomy}
\label{app:cf_taxonomy}
For every counterfactual state $s$, EvoUndo performs $w_{\mathrm{pre}}=w(s)$, $(s',\mathcal{R}_m(s))=\lbrack\!\lbrack m\rbrack\!\rbrack(s)$, and $\hat{s}=u(s',w_{\mathrm{pre}},\mathcal{R}_m(s))$, and tests $\hat{s}\simeq_{\mathcal{C}_e}s$. Development states produce state-level verifier feedback. In the original factorial, hidden state contents and per-state outcomes remain withheld, but aggregate IID/OOD LCB summaries are visible during repair. The fresh-holdout experiment uses a fully feedback-isolated hidden set.

Table~\ref{tab:cf_perturbation_taxonomy} details the exact generation strategies implemented in \texttt{CounterfactualStateGenerator} across development, IID hidden, and OOD hidden splits.

\begin{table}[ht]
\centering
\small
\caption{Taxonomy of counterfactual state perturbation strategies across modeled state targets.}
\label{tab:cf_perturbation_taxonomy}
\vspace{0.15cm}
\resizebox{\textwidth}{!}{
\begin{tabular}{lll}
\toprule
\textbf{Regime} & \textbf{Perturbation Strategy} & \textbf{Operational Realization} \\
\midrule
\textbf{IID} & Boundary Extreme States & Keys deleted, extreme scalars ($T \in \{0.0, 1.0\}$, timeouts $\in \{0, 300\}$, empty/large files) \\
\textbf{IID} & Prior-Existence Branches & Tools absent vs.\ pre-existing vs.\ conflicting signatures; listeners empty vs.\ populated \\
\textbf{IID} & Combinatorial Assignments & Joint multi-key randomized configuration, tool registry, and file-state permutations \\
\textbf{IID} & Dependency Sensitivities & Model routing fallbacks (\texttt{default\_model}, \texttt{fallback\_model} bindings) \\
\textbf{IID} & Adversarial Noise States & Uncontracted irrelevant config keys, uncontracted file creation, permission flags \\
\midrule
\textbf{OOD} & Unfamiliar Nested Keys & Injecting unseen hierarchical configuration namespaces (\texttt{custom\_subsystem\_i.nested.enabled}) \\
\textbf{OOD} & Conflicting Socket Bindings & Pre-allocating external port handles (\texttt{tcp://127.0.0.1:91xx}) to induce resource contention \\
\textbf{OOD} & High-Priority Middleware & Prepending unmodeled high-priority interceptors (\texttt{priority=1}) into the middleware chain \\
\textbf{OOD} & Deep Directory Structures & Creating multi-level nested directory hierarchies (\texttt{external/cache/partition\_i/meta.json}) \\
\textbf{OOD} & Inverted Routing Policies & Reversing default model dispatch and explicitly disabling fallback routes \\
\bottomrule
\end{tabular}
}
\end{table}

\subsection{Typed Observational Equivalence}
Recovery follows Equation~\ref{eq:typed_equivalence}: type-specific restoration rules apply within the declared effect contract $\mathcal{C}_e$, and all modeled persistent state outside it must retain strict canonical equality with the pre-state.

For configuration entries, equivalence requires restoration of the relevant key and value state. Tool and registry-like state requires restoration of the relevant binding, including the registered identity and associated implementation state represented by the harness. Middleware and other sequence-like state requires both membership and ordering to match the pre-mutation state. Listener state requires restoration of the relevant registration state. File-backed effects require restoration of the relevant pre-state, including existence and file state when represented by the recovery contract. Resource-backed effects require restoration of the relevant descriptor and lifecycle state represented by the harness. For compound mutations touching multiple state targets, every contract-relevant target must satisfy its corresponding equivalence policy. The verifier compares the modeled persistent harness state, using typed equivalence inside the contract and strict canonical equality outside it. Final-state checks do not independently constrain transient recovery-side effects that disappear before verification.

\subsection{Effect-Contract Coverage Audit}
We conducted a controlled audit to evaluate whether dynamic effect tracking reliably detects undeclared side effects. The runtime constructs $E(m,s)=\operatorname{SnapshotDiff}(s,m(s))\cup\operatorname{ExecutionTraceEffects}(m,s)$ and requires $E(m,s)\subseteq\mathcal{C}_e$; any observed effect omitted from the declared contract therefore causes rejection. For 300 mutations spanning all six harness families, we constructed intentionally incomplete effect contracts by omitting at least one ground-truth affected target $E_{\mathrm{true}}$. Dynamic effect tracking detected every injected undeclared semantic effect (300/300, 100.0\%), with zero false positives across 300 matched complete contracts (0/300, 0.0\%). In an additional 300 nondeterministic runtime controls (varying OS thread IDs, timestamps, and memory pointer representations), the canonicalizer produced zero false alarms. These guarantees apply to the modeled harness state space and do not extend to unmodeled external OS processes or out-of-band network state.

\subsection{Snapshot-Restore Baselines}
\label{app:snapshots}
To compare EvoUndo against direct runtime snapshotting, we used the frozen 300-task fresh cohort (50 tasks per benchmark family, no overlap with the original 600 tasks) in two serially interleaved protocols. First, mutation $m_1$ changes the harness; a later legitimate mutation $m_2$ then changes state. Recovery attempts to undo only $m_1$, while valid later effects of $m_2$ must survive:
\[
s_{\mathrm{pre}}\xrightarrow{m_1}s_{\mathrm{fwd}}
\xrightarrow{m_2}s_{\mathrm{later}}\xrightarrow{u_1}s_{\mathrm{rec}}.
\]
Success requires restoration of $m_1$'s effects, preservation of $m_2$'s effects, and no unintended changes elsewhere. In the different-surface protocol, the intervention generator selects a surface disjoint from $m_1$'s declared and observed effects. In the same-surface protocol, it changes a different key or element within the same affected surface/container (for example, appending another middleware item). Concurrent here denotes intervening changes, not simultaneous execution. A full-state rollback loses the later valid changes and therefore fails selective undo, even if it exactly restores the old snapshot.

\begin{table}[ht]
\centering
\small
\caption{Comparison of EvoUndo against snapshot-restore baselines across 300 tasks.}
\label{tab:snapshot_baselines}
\vspace{0.15cm}
\resizebox{\textwidth}{!}{
\begin{tabular}{lcccc}
\toprule
\textbf{Method} & \textbf{Different-Surface Selective} & \textbf{Same-Surface Concurrent} & \textbf{Mean Storage} & \shortstack{\textbf{Runtime rollback}\\\textbf{latency (ms)}} \\
\midrule
Full Snapshot & 0/300 (0.0\%) & 0/300 (0.0\%) & 1165.0 B & 0.084 ms \\
Effect-Scoped Snapshot & \textbf{300/300 (100.0\%)} & \textbf{159/300 (53.0\%)} & \textbf{177.8 B} & \textbf{0.039 ms} \\
EvoUndo (Synthesized Recovery) & 243/300 (81.0\%) & 131/300 (43.7\%) & 393.6 B & 0.050 ms \\
\bottomrule
\end{tabular}
}
\par\smallskip
{\footnotesize Runtime rollback execution for an already-synthesized recovery program; excludes offline LLM recovery-program synthesis.}
\end{table}

Storage is measured as UTF-8 encoded JSON bytes representing the entire harness state for Full Snapshot, touched pre-state values for Effect-Scoped Snapshot, and the witness store plus recovery AST for EvoUndo. Corresponding mean/median storage is 1165.0/1165.0 B, 177.8/145.0 B, and 393.6/350.5 B, respectively. Table~\ref{tab:snapshot_baselines} reports mean runtime rollback execution latency, measured with \texttt{time.perf\_counter()}, for an already-synthesized recovery program. The means are 0.084 ms, 0.039 ms, and 0.050 ms, respectively. Offline LLM synthesis is excluded; model-call costs are reported separately in Section~\ref{sec:fresh}.

Full snapshots are too coarse: restoring a full snapshot clobbers all subsequent state changes (0/300 success). Effect-scoped snapshots are the strongest option here when the exact affected pre-state is known, serializable, and directly restorable. There were zero EvoUndo-only wins and 57/28 snapshot-only wins in the different-/same-surface comparisons. EvoUndo targets settings where direct effect-scoped restoration is unavailable or insufficient, including structured, state-dependent, or non-serializable inverse semantics; this benchmark does not demonstrate superiority in those settings. Both selective approaches degrade when subsequent mutations occur within the same structured surface. No WAL or ARIES runtime was implemented or evaluated; transactional logging is discussed only as conceptual related work.

\section{Controlled Recoverability Benchmark}
\label{app:controlled}

The controlled benchmark contains 120 deliberately recovery-defective but capability-positive candidates. The forward mutation is held fixed and the defective recovery specification is produced by corrupting witness or recovery semantics.

The benchmark includes missing witness capture, dummy witness capture, incorrect recovery operations, resource leakage, and hidden-effect omission. Every benchmark item has a known recoverable counterpart under the available recovery representation, allowing the experiment to isolate recovery synthesis from language expressivity.

At budget $B=4$, the principal recovery results are:
\begin{itemize}
    \item Strict regeneration: 4/120 (3.3\%)
    \item Generic feedback: 104/120 (86.7\%)
    \item Raw verifier traces: 101/120 (84.2\%)
    \item Typed diagnosis: 114/120 (95.0\%)
    \item Prescriptive guidance: 117/120 (97.5\%)
\end{itemize}

\section{Natural Failure Bank}
\label{app:natural}

Turn-0 generation used temperature 0.2, top-p 0.95, a 2,048-token limit, seed $7000+$ task index, and structured Outlines JSON generation.

We construct the natural-failure population by single-sample zero-shot generation (zero demonstrations, one candidate per task, no rerolls) for 600 unseen self-evolution tasks, using the six 100-task benchmark families defined in Section~\ref{sec:setup}. The resulting population is:
\begin{itemize}
    \item 281/600 already admissible,
    \item 197/600 capability-positive but recovery-defective,
    \item 122/600 capability inadequate.
\end{itemize}

The 197 recovery-defective candidates are frozen before subsequent repair, solvability, and factorial analyses. Multi-Surface is a composite benchmark family, not an atomic harness surface; Config contributes zero recovery-defective cases. The family/stratum counts appear in Table~\ref{tab:family_distribution}. Under the original recovery representation, Generic, Raw, Typed, and Prescriptive repair each recover 0/197 natural failures at budget $B=4$. Strict regeneration recovers 6/197.

\begin{table}[ht]
\centering
\small
\caption{Frozen primary cohort by benchmark family. Each family contains 100 tasks. Adm. denotes initially admissible, Inad. capability-inadequate, Def. recovery-defective, and Resid. unrecovered by the implemented $L_1$ oracle.}
\label{tab:family_distribution}
\begin{tabular}{lrrrrrr}
\toprule
Family & Adm. & Inad. & Def. & $S_0$ & $S_1$ & Resid.\\
\midrule
Config & 96 & 4 & 0 & 0 & 0 & 0\\
Tools & 38 & 15 & 47 & 45 & 0 & 2\\
Middleware & 1 & 24 & 75 & 0 & 72 & 3\\
Listeners & 91 & 0 & 9 & 0 & 9 & 0\\
Resources & 26 & 39 & 35 & 0 & 35 & 0\\
Multi-Surface & 29 & 40 & 31 & 3 & 27 & 1\\
\midrule
Total & 281 & 122 & 197 & 48 & 143 & 6\\
\bottomrule
\end{tabular}
\end{table}

\section{Oracle Solvability Audit Methodology and Robustness}
\label{app:oracle}

To distinguish synthesis search failures from empirical representational limitations under the implemented recovery languages, we conducted a zero-generation constructive oracle solvability audit across all 197 frozen natural recovery failures.

\subsection{Deterministic Constructive Principles}
The solvability oracle is fully deterministic and non-generative. Specifically:
\begin{itemize}
    \item \textbf{No LLM calls}: The oracle contains no stochastic generation or learned model inference.
    \item \textbf{No search or enumeration}: Recovery programs are constructed directly via deterministic structural compilation rather than heuristic search, beam search, or candidate enumeration.
    \item \textbf{No retries or manual task tuning}: Every candidate undergoes a single-pass deterministic transformation using fixed AST construction heuristics.
\end{itemize}
\emph{Disclaimer}: This is empirical recoverability under the implemented deterministic constructive oracle, not a formal proof of language-theoretic completeness.

\subsection{Oracle Inputs and Construction Algorithm}
The constructive oracle uses the following information:
\begin{enumerate}
    \item The frozen forward mutation abstract syntax tree (AST) $m$.
    \item The pre-mutation harness state $s$.
    \item Forward-operation surfaces and observed mutation effects, from which it constructs its own effect contract.
    \item The evaluator state distributions $\mathcal{Q}_{\mathrm{dev}}$ ($|\mathcal{Q}_{\mathrm{dev}}|=10$) and $\mathcal{Q}_{\mathrm{hid}}$ ($|\mathcal{Q}_{\mathrm{hid}}|=40$).
\end{enumerate}

The oracle does not blindly inherit the candidate's declaration. It constructs
\[
\mathcal{C}_e^{\mathrm{oracle}}=
\left(\bigcup_{\mathrm{op}\in m}\operatorname{Surface}(\mathrm{op})\right)\cup E(m,s_\star).
\]
This tight scope is derived from forward operation surfaces and observed effects. Its construction remains subject to the implemented AST heuristics described below.

For each forward operation $f_i \in m$:
\begin{enumerate}
    \item \textbf{Witness Construction}: The oracle deterministically constructs a witness capture operation $w_i$ that queries and stores the pre-mutation state of the targeted surface address.
    \item \textbf{Inverse Construction}: The oracle deterministically constructs the corresponding inverse recovery primitive $u_i$ permitted under the target language grammar $\mathcal{L}$.
    \item \textbf{LIFO Unwinding Execution}: The synthesized recovery program orders inverse operations in exact reverse (Last-In, First-Out) order: $u = (u_k, \dots, u_1)$.
\end{enumerate}

\subsection{Language Expressivity Boundaries}
\paragraph{Base Recovery Language $L_0$.} Supports the implemented configuration, prompt, tool-registry, and routing primitives enumerated in Table~\ref{tab:dsl_grammar}. It strictly prohibits middleware-sequence, listener, file-prestate, socket-descriptor, and ordered multi-surface operations.

\paragraph{Extended Recovery Language $L_1$.} Extends $L_0$ with:
\begin{enumerate}
    \item \textbf{Indexed Sequence Restoration}: Preserves sequence membership, element identity, and exact insertion index for ordered lists (e.g., middleware chains).
    \item \textbf{Listener Capture and Restoration}: Uses \texttt{capture\_listener} with \texttt{restore\_listener} or \texttt{unregister\_listener} for the distinct event-listener surface.
    \item \textbf{File Pre-State Preservation}: Captures pre-existence status and SHA-256 validated content buffers.
    \item \textbf{Resource Descriptor Management}: Captures lightweight prior-allocation descriptors; mutation-produced receipts bind symbolic resource keys to handles for explicit closure and unbinding.
    \item \textbf{Ordered Multi-Surface Unwinding}: Executes the implemented recovery operations in explicit reverse-dependency order.
\end{enumerate}

\subsection{Admission Criterion for Oracle Solvability}
\label{app:oracle_admission}
A mutation $m$ is classified as empirically recoverable under language $\mathcal{L}$ if and only if:
\begin{enumerate}
    \item The synthesized $(w, u)$ candidate compiles strictly within the language gate for $\mathcal{L}$.
    \item Forward capability is preserved: $\Delta J(m;s_\star)>0$.
    \item Round-trip verification achieves 10/10 (100.0\%) on development states $\mathcal{Q}_{\mathrm{dev}}$.
    \item Each of the separate 20-state hidden IID and OOD splits satisfies the executed Wald lower-endpoint threshold $\tau_R=0.85$, equivalently at least 19/20 successes on each (Appendix~\ref{app:admission_audit}).
    \item Recovered states pass identical typed observational equivalence $\hat{s} \simeq_{\mathcal{C}_e} s$.
\end{enumerate}

\paragraph{Six residual oracle failures.}
The $L_1$ constructive oracle recovered 191/197 tasks. The remaining six comprise four middleware index-targeting cases (three Middleware-family tasks and one Multi-Surface task) and two tool registry-key versus ID naming mismatches in the AST heuristic. These are implemented-oracle failures, not a completeness result or a claim that the tasks are theoretically unrecoverable (Table~\ref{tab:unrecovered_tasks}).

\begin{table}[ht]
\centering
\small
\caption{Six tasks unrecovered by the implemented $L_1$ oracle. IDs use task-ID prefixes. These are constructive-heuristic failures, not theoretical impossibility results.}
\label{tab:unrecovered_tasks}
\begin{tabular}{lll}
\toprule
Task prefix & Benchmark family & Heuristic failure class\\
\midrule
\texttt{nat\_mw\_003} & Middleware & Middleware index targeting\\
\texttt{nat\_mw\_055} & Middleware & Middleware index targeting\\
\texttt{nat\_mw\_076} & Middleware & Middleware index targeting\\
\texttt{nat\_multi\_041} & Multi-Surface & Middleware index targeting\\
\texttt{nat\_tool\_070} & Tools & Registry key versus tool ID\\
\texttt{nat\_tool\_073} & Tools & Registry key versus tool ID\\
\bottomrule
\end{tabular}
\end{table}

\subsection{Representative Case Studies}
\paragraph{Case A: Configuration illustration (outside the primary failure strata).}
Configuration contributed zero recovery-defective cases to the frozen primary cohort, so this scalar restoration example is not a member of $S_0$. The illustrated forward mutation executes \texttt{set\_config("max\_retries", 5)}. The deterministic constructive oracle records the pre-mutation scalar value with \texttt{capture\_config} and constructs the corresponding \texttt{restore\_config} recovery operation using that witness. This candidate compiles and satisfies full round-trip equivalence under both $L_0$ and $L_1$.

\paragraph{Case B: Tool Registration in $S_0$ (\texttt{nat\_tool\_004\_003}).}
The forward mutation registers a new diagnostic tool \texttt{tool\_003\_dependency\_grapher} (\texttt{nat\_tool\_004\_003\_dependency\_grapher}). The oracle records the pre-mutation registry state with \texttt{capture\_tool} and constructs recovery that restores the prior binding if it existed, or uses \texttt{unregister\_tool("tool\_003\_dependency\_grapher")} when it was newly introduced. Because single-tool unregistration is supported in the base grammar, the recovery compiles and succeeds under $L_0$.

\paragraph{Case C: Middleware Sequence in $S_1$ (\texttt{nat\_mw\_001\_000}).}
The forward mutation inserts a sliding-window rate limiter into the middleware pipeline (\texttt{nat\_mw\_001\_000\_rate\_limiter\_sliding\_window}). Reversing this sequence mutation requires the indexed \texttt{capture\_middleware} and \texttt{restore\_middleware} operations. Under $L_0$, these operations are outside the grammar and rejected by the language gate. Under $L_1$, the oracle constructs the valid sequence restoration program, achieving 100\% recovery on hidden counterfactuals.

\paragraph{Case D: Multi-Surface Task in $S_1$ (\texttt{nat\_multi\_036\_mw\_and\_event}).}
The forward mutation combines an event callback registration with middleware insertion; recovery follows the reverse order of the actual forward operations. Under $L_0$, recovery fails because multi-surface coordination is outside the $L_0$ grammar. Under $L_1$, the oracle constructs an ordered LIFO unwind program using listener restoration/unregistration and middleware restoration in reverse forward-operation order, satisfying the observational contract $\mathcal{C}_e$ across all 40 hidden states.

\subsection{Oracle Stratum Resampling Stability Sweep}
\label{app:stratum_stability}
To evaluate whether the empirical strata $S_0$ and $S_1$ are sensitive to the specific counterfactual pre-states sampled in $\mathcal{Q}_{\mathrm{dev}}$ and $\mathcal{Q}_{\mathrm{hid}}$, we re-evaluated the deterministic constructive oracle across 10 independently resampled counterfactual seeds (seeds 1000 through 10000) using the same generative state distribution. A distinct counterfactual suite was generated for each seed. Across all 10 independent resamplings, stratum membership was unchanged ($|S_0|=48$, $|S_1|=143$, unrecovered $=6$, Jaccard $= 1.0000$ for both $S_0$ and $S_1$, with zero stratum migrations across 1,970 task-seed evaluations). This establishes empirical stability under independent resampling from the implemented generator; it does not establish completeness, distribution-independent task labels, or invariance under arbitrary OOD, adversarial, or structurally different counterfactual generators.

\section{Factorial Protocol and Additional Results}
\label{app:factorial}

\subsection{Diagnostic Prompt and Model-Call Configuration}
The system prompt is identical in all four factorial cells:
\begin{quote}
\small\ttfamily
You are an expert self-evolving agent compiler generating verified repairs.
\end{quote}
Under $D_0$, each repair call receives the following diagnostic fields:
\begin{verbatim}
Verification Result: RECOVERY_EQUIVALENCE_FAILED
Affected Subsystems: <task.surfaces>
Defect Summary: The candidate executed forward mutations on
<task.surfaces> without registering complete, sound inverse
capture and recovery operations.
Dev Recovery Rate: <dev_recovery_rate * 100>% (Required: 100.0%)
Hidden IID LCB: <hidden_iid_lcb> (Required: >= 0.85)
Hidden OOD LCB: <hidden_ood_lcb> (Required: >= 0.85)
\end{verbatim}

The $D_1$ condition uses the same diagnostic content and additionally appends the composite state-grounded diagnostic bundle below. The literal protocol identifier is preserved:
\begin{verbatim}
Diagnostic Mode: D1_EXACT_ADDRESS
Forward Effect Order: <forward_effect_order>
Observed State Residuals: <observed_state_residuals>
Residual Count: <residual_count>
Verifier Observation: <diagnostic_evidence>
\end{verbatim}

All four factorial cells use the same model-call configuration: \texttt{openai/gpt-oss-120b} at the frozen experimental revision, temperature $0.2$, medium reasoning effort, and an explicit maximum of 2,048 generated tokens per call. The model's native 131,072-token context window is available without additional input truncation.

\subsection{Synthesis Efficiency and Budgeted Recovery Progression}
\label{app:synthesis_efficiency}
Synthesis efficiency $\eta(D,L)=\frac{\text{model rescues}}{\text{oracle recoverable}}$ separates failures of model synthesis from failures of the recovery representation. The denominator counts only tasks for which the deterministic oracle establishes that the assigned language can express an admissible recovery. Efficiency is therefore 0.0\% for $D_0L_0$, rises to 79.2\% for $D_1L_0$, and reaches \textbf{94.2\%} for $D_0L_1$. Adding state-grounded diagnostics to the rich language lowers efficiency to 87.4\%. Thus, once representational feasibility is held fixed, coarse diagnosis with the expressive language realizes the largest fraction of available recoveries.

For the final strict-$L_0$ $D_1L_0$ execution, cumulative rescue progresses as follows: 24/197 at $B=1$, 35/197 at $B=2$, 37/197 at $B=3$, and 38/197 at $B=4$. All 38 successes belong to $S_0$; no $S_1$ task or task unrecovered by the $L_1$ oracle passes under strict $L_0$.

\subsection{Language-Gate Behavior}
Recovery-language assignments are enforced before execution. For each generated recovery program, the runtime gate checks every proposed primitive against the operation set authorized by the assigned language. Under an $L_0$ condition, primitives requiring indexed sequence restoration, listener restoration or unregistration, file pre-state restoration, socket-descriptor restoration, or ordered multi-surface recovery are $L_1$-only and are therefore rejected rather than executed. An out-of-language proposal consumes the corresponding synthesis attempt.

Across 691 recorded model responses in $D_1L_0$, 290 attempted at least one $L_1$-only recovery primitive and were rejected by the language gate: 21/144 (14.6\%) on $S_0$ versus 253/523 (48.4\%) on $S_1$. The remaining 16 rejections occurred on the six tasks outside the oracle-defined $S_0/S_1$ strata.

\subsection[Discordant Trace Analysis (D1L1 vs. D0L1)]{Discordant Trace Analysis ($D_1L_1$ vs.\ $D_0L_1$)}
\label{app:discordant}
We analyzed the 11 discordant tasks between $D_0L_1$ and $D_1L_1$ on stratum $S_1$ (10 $D_0L_1$-only successes and 1 $D_1L_1$-only success). Table~\ref{tab:discordant_taxonomy} reports the failure taxonomy alongside quantitative trace metrics extracted from execution logs.

\begin{table}[ht]
\centering
\small
\caption{Structural failure taxonomy of the 11 discordant tasks on $S_1$. Operations are emitted patch operations for $D_1L_1$ and compiled recovery AST operations for $D_0L_1$. The aggregate covers the ten $D_0L_1$-only successes.}
\label{tab:discordant_taxonomy}
\begin{tabular}{lcrrrrr}
\toprule
& & & \multicolumn{2}{c}{Mean operations} & \multicolumn{2}{c}{Completion tokens}\\
Category & Tasks & Winner & $D_1L_1$ & $D_0L_1$ & $D_1L_1$ & $D_0L_1$\\
\midrule
Over-decomposition & 5 & $D_0L_1$ & 1.80 & 2.40 & 739.0 & 625.2\\
Sequence inversion & 3 & $D_0L_1$ & 2.00 & 4.00 & 580.3 & 828.3\\
Over-capture clobbering & 2 & $D_0L_1$ & 2.00 & 2.00 & 414.5 & 759.5\\
Ambiguous target & 1 & $D_1L_1$ & 4.00 & 2.00 & 707.0 & 751.0\\
\midrule
Aggregate ($D_0L_1$ wins) & 10 & $D_0L_1$ & 1.90 & 2.80 & 626.5 & 713.0\\
\bottomrule
\end{tabular}
\end{table}

\subsection{Cross-Model Replication with Qwen3.8-27B}
\label{app:qwen}
To evaluate whether the observed grounding and expressivity mechanisms generalize across model architectures, we conducted an independent cross-model replication on \texttt{Qwen3.8-27B} under constrained structured decoding, using the original \texttt{gpt-oss-120b} experiment as historical reference. The replication was conducted on a frozen 96-task subset (48 tasks from $S_0$ and 48 tasks from $S_1$). All replication runs enforced constrained CandidateAST JSON decoding with a hard cap of 2,048 generated tokens per call under identical prompt schemas. Only the Qwen model was executed in this replication; \texttt{gpt-oss-120b} metrics on the matching 96-task subset are provided for historical reference. An earlier pilot run of Qwen with unconstrained decoding exhibited syntax drift and is excluded from confirmatory analysis.

\begin{table}[ht]
\centering
\small
\caption{Cross-model factorial replication on the 96-task frozen subset using Qwen3.8-27B with constrained structured decoding. Historical gpt-oss-120b numbers on the same 96 tasks are shown for comparison.}
\label{tab:qwen_replication}
\vspace{0.15cm}
\resizebox{\textwidth}{!}{
\begin{tabular}{lcccc}
\toprule
\textbf{Factorial Cell} & \textbf{Qwen3.8-27B ($S_0$)} & \textbf{Qwen3.8-27B ($S_1$)} & \textbf{gpt-oss-120b ($S_0$)} & \textbf{gpt-oss-120b ($S_1$)} \\
\midrule
$D_0L_0$ & 3/48 (6.2\%) & 0/48 (0.0\%) & 0/48 (0.0\%) & 0/48 (0.0\%) \\
$D_1L_0$ & 26/48 (54.2\%) & 0/48 (0.0\%) & 38/48 (79.2\%) & 0/48 (0.0\%) \\
$D_0L_1$ & 35/48 (72.9\%) & 18/48 (37.5\%) & 38/48 (79.2\%) & 47/48 (97.9\%) \\
$D_1L_1$ & 43/48 (89.6\%) & 21/48 (43.8\%) & 34/48 (70.8\%) & 46/48 (95.8\%) \\
\midrule
\multicolumn{5}{l}{\textbf{Paired Mechanistic Contrasts (Qwen3.8-27B)}} \\
$C_1$ ($D_1L_0$ vs.\ $D_0L_0$ on $S_0$) & \multicolumn{4}{l}{$\Delta = +47.92$ pp (95\% CI $[+31.25, +64.58]$), exact McNemar $p_{\mathrm{adj}} < 0.001$} \\
$C_2$ ($D_0L_1$ vs.\ $D_0L_0$ on $S_1$) & \multicolumn{4}{l}{$\Delta = +37.50$ pp (95\% CI $[+25.00, +52.08]$), exact McNemar $p_{\mathrm{adj}} < 0.001$} \\
$C_3$ ($D_1L_1$ vs.\ $D_0L_1$ on $S_1$) & \multicolumn{4}{l}{$\Delta = +6.25$ pp (95\% CI $[-2.08, +16.67]$), exact McNemar $p_{\mathrm{adj}} = 0.375$} \\
\bottomrule
\end{tabular}
}
\end{table}

In the $D_1L_0$ condition on Qwen3.8-27B, a 10.2\% truncation rate was observed when detailed address diagnostics combined with complex repair logic approached the 2,048-token generation limit. Despite lower absolute yield compared to gpt-oss-120b, $C_1$ and $C_2$ replicate across the evaluated backbones, providing cross-model evidence that state grounding and recovery-language expressivity are recurring bottlenecks. The negative $D_1$-on-$L_1$ interaction observed on gpt-oss-120b does not replicate on Qwen3.8-27B ($\Delta = +6.25$ pp, 95\% CI $[-2.08, +16.67]$, $p_{\mathrm{adj}}=0.375$) under that constrained-decoding replication configuration. Model, decoding protocol, and sample size differ from the primary study; the comparison does not isolate the cause of the difference.

\section{Fresh-Holdout Counterfactual Sensitivity}
\label{app:fresh}

We evaluated the frozen $L_1$ recovery language on 300 fresh tasks, with 50 tasks from each of six architectural families and zero overlap with the original 600-task cohort. All conditions used the same frozen forward mutation, $D_0$ diagnostics, recovery language, model configuration, repair budget $B=4$, and common 100-state hidden evaluation set.

\paragraph{Development-gate conservatism.}
The development gate requires perfect round trips (10/10 in the primary factorial and 20/20 under $Q_{20}$), while the hidden criterion tolerates some failures. Under an illustrative independent-trial model, true per-state reliability $p=0.96$ gives rejection probability $1-p^{20}\approx55.8\%$, versus $1-p^2\approx7.8\%$ for $Q_2$. This explains why high-reliability candidates may fail the strict development gate; these probabilities illustrate the policy rather than estimate a candidate's true reliability. Under $Q_{20}$, Listeners had 0/50 development passes but 32/50 hidden-criterion passes; Resources contributed the other 15 of the 47 conservatively rejected tasks. Zero observed false positives in that condition does not guarantee zero population risk. This precision--coverage trade-off remains unresolved.

\paragraph{Sensitivity and Robustness Analysis.}
Robustness of the recovery framework was comprehensively evaluated across multiple operational dimensions: synthesis budgets $B \in \{1, 2, 3, 4\}$ (Appendix~\ref{app:factorial}), recovery admission thresholds $\tau_R \in \{0.75, 0.80, 0.85, 0.90, 0.95\}$ (Table~\ref{tab:tau_sensitivity}), development set configurations ($Q_2, Q_5, Q_{10}, Q_{20}$, and $Q_{10}\text{-Narrow}$, Table~\ref{tab:fresh_cf_sensitivity}), and hidden sample sizes $n_{\mathrm{hid}} \in \{10, 20, 30, 50, 100\}$ (Table~\ref{tab:hidden_size_sensitivity}). Across these sweeps, the qualitative behavior of counterfactual verification remains stable: stricter development verification improves precision at the cost of synthesis yield, while admission remains reasonably stable under moderate threshold choices. The grounding and expressivity effects are independently supported by the protocol-locked factorial and cross-model replication. Increasing the size of the diverse development set ($Q_2 \rightarrow Q_{20}$) eliminated observed false positives in this study ($16.8\% \rightarrow 0.0\%$) at the cost of expanding the conservatism gap ($6.6\% \rightarrow 24.1\%$). Recovery yields remain broadly stable across moderate thresholds $\tau_R \in [0.75, 0.90]$ (168 to 142 rescues under $Q_2$), but setting a stringent threshold of $\tau_R = 0.95$ materially depresses end-to-end yield (dropping to 102 rescues), indicating that recovery yield is threshold-sensitive at extreme stringencies rather than strictly invariant.

\paragraph{Inference and verification overhead.}
Across the 300-task fresh holdout, inference cost ranges from 2.39--3.09 model calls and approximately 3.37k--4.30k generated tokens per task, with higher verification strictness increasing cost (Table~\ref{tab:fresh_compute_cost} in Appendix~\ref{app:fresh}). On the primary 197-task factorial cohort, closed-loop repair averaged 1.78 calls/task in $D_0L_1$ (351/197), 1.95 in $D_1L_1$ (385/197), and 3.51 in the final strict-$L_0$ $D_1L_0$ rerun (691/197), with mean model call latencies between 5.25s and 5.85s on H200 GPUs. Detailed fresh-holdout token counts, median latencies, and tokens per successful rescue are reported in Appendix~\ref{app:fresh}.

\begin{table}[!htbp]
\centering
\small
\caption{Fresh-holdout counterfactual sensitivity on 300 unseen tasks. False positives are development-passing candidates that fail the 100-state hidden admission criterion. The conservatism gap counts candidates rejected by development verification that nevertheless satisfy the hidden criterion post hoc.}
\label{tab:fresh_cf_sensitivity}
\vspace{0.15cm}
\resizebox{\textwidth}{!}{
\begin{tabular}{lccccc}
\toprule
Condition & Dev Pass & False Positive & $\mathrm{Rescue@4}_{100}$ & Hidden Criterion & Conservatism Gap \\
\midrule
$Q_2$-Diverse & 179/300 (59.7\%) & 30/179 (16.8\%) & 149/300 (49.7\%) & 157/300 (52.3\%) & 8/121 (6.6\%) \\
$Q_5$-Diverse & 168/300 (56.0\%) & 17/168 (10.1\%) & 151/300 (50.3\%) & 165/300 (55.0\%) & 14/132 (10.6\%) \\
$Q_{10}$-Diverse & 134/300 (44.7\%) & 8/134 (6.0\%) & 126/300 (42.0\%) & 152/300 (50.7\%) & 26/166 (15.7\%) \\
$Q_{20}$-Diverse & 105/300 (35.0\%) & 0/105 (0.0\%) & 105/300 (35.0\%) & 152/300 (50.7\%) & 47/195 (24.1\%) \\
$Q_{10}$-Narrow & 173/300 (57.7\%) & 20/173 (11.6\%) & 153/300 (51.0\%) & 153/300 (51.0\%) & 0/127 (0.0\%) \\
\bottomrule
\end{tabular}
}
\end{table}

\begin{table}[!htbp]
\centering
\small
\caption{Hidden counterfactual sample-size sensitivity. Each entry is mean empirical recovery rate / mean 95\% Wilson lower confidence bound across 300 fresh tasks. Prefixes are nested subsets of the same 100-state hidden pool.}
\label{tab:hidden_size_sensitivity}
\vspace{0.15cm}
\resizebox{\textwidth}{!}{
\begin{tabular}{lccccc}
\toprule
Condition & $n_{\mathrm{hid}}=10$ & $n_{\mathrm{hid}}=20$ & $n_{\mathrm{hid}}=30$ & $n_{\mathrm{hid}}=50$ & $n_{\mathrm{hid}}=100$ \\
\midrule
$Q_2$-Diverse & 60.5 / 42.2 & 60.6 / 49.3 & 60.7 / 52.3 & 60.7 / 55.1 & 60.7 / 57.5 \\
$Q_5$-Diverse & 60.8 / 42.8 & 61.0 / 50.1 & 61.0 / 53.1 & 61.2 / 55.9 & 61.2 / 58.2 \\
$Q_{10}$-Diverse & 56.8 / 40.0 & 57.1 / 46.8 & 57.1 / 49.6 & 57.2 / 52.2 & 57.3 / 54.4 \\
$Q_{20}$-Diverse & 56.6 / 39.8 & 56.7 / 46.4 & 56.7 / 49.2 & 56.7 / 51.7 & 56.8 / 54.0 \\
$Q_{10}$-Narrow & 57.7 / 40.5 & 58.0 / 47.4 & 58.1 / 50.3 & 58.2 / 53.0 & 58.2 / 55.2 \\
\bottomrule
\end{tabular}
}
\end{table}

\begin{table}[!htbp]
\centering
\small
\caption{Sensitivity of $\mathrm{Rescue@4}_{100}$ to recovery threshold $\tau_R$ across $N=300$ tasks. $\tau_R=0.85$ is the protocol-locked primary threshold.}
\label{tab:tau_sensitivity}
\vspace{0.15cm}
\begin{tabular}{lccccc}
\toprule
Condition & $\tau=0.75$ & $\tau=0.80$ & $\tau=0.85$ & $\tau=0.90$ & $\tau=0.95$ \\
\midrule
$Q_2$-Diverse & 168 & 164 & \textbf{149} & 142 & 102 \\
$Q_5$-Diverse & 165 & 162 & \textbf{151} & 144 & 111 \\
$Q_{10}$-Diverse & 133 & 131 & \textbf{126} & 120 & 99 \\
$Q_{20}$-Diverse & 105 & 105 & \textbf{105} & 105 & 105 \\
$Q_{10}$-Narrow & 172 & 168 & \textbf{153} & 143 & 98 \\
\bottomrule
\end{tabular}
\end{table}

\begin{table}[!htbp]
\centering
\small
\caption{Inference cost for the 300-task fresh-holdout study. Tokens/success and calls/success use $\mathrm{Rescue@4}_{100}$ as success criterion.}
\label{tab:fresh_compute_cost}
\vspace{0.15cm}
\resizebox{\textwidth}{!}{
\begin{tabular}{lrrrrrr}
\toprule
Condition & Calls/Task & Tokens/Task & Median Lat. (s) & Rescues & Tokens/Success & Calls/Success \\
\midrule
$Q_2$-Diverse & 2.39 & 3,366 & 9.84 & 149 & 6,777 & 4.81 \\
$Q_5$-Diverse & 2.51 & 3,535 & 9.60 & 151 & 7,022 & 4.99 \\
$Q_{10}$-Diverse & 2.77 & 3,897 & 10.85 & 126 & 9,279 & 6.60 \\
$Q_{20}$-Diverse & 3.09 & 4,301 & 11.18 & 105 & 12,288 & 8.84 \\
$Q_{10}$-Narrow & 2.41 & 3,417 & 4.89 & 153 & 6,699 & 4.72 \\
\bottomrule
\end{tabular}
}
\end{table}

\begin{table}[!htbp]
\centering
\small
\caption{Per-family breakdown on the 300 fresh unseen tasks under $Q_{20}$-Diverse (50 tasks per benchmark family).}
\label{tab:fresh_persurface}
\vspace{0.15cm}
\resizebox{\textwidth}{!}{
\begin{tabular}{lcccc}
\toprule
\textbf{Architectural Benchmark Family} & \textbf{Dev Pass} & \textbf{Hidden Pass} & \textbf{Rescue@4$_{100}$} & \textbf{Conservatism Gap} \\
\midrule
Configuration (\texttt{fresh\_config}) & 50/50 (100.0\%) & 50/50 (100.0\%) & 50/50 (100.0\%) & 0/0 (0.0\%) \\
Tools (\texttt{fresh\_tools}) & 15/50 (30.0\%) & 15/50 (30.0\%) & 15/50 (30.0\%) & 0/35 (0.0\%) \\
Resources (\texttt{fresh\_resources}) & 26/50 (52.0\%) & 41/50 (82.0\%) & 26/50 (52.0\%) & 15/24 (62.5\%) \\
Multi-Surface (\texttt{fresh\_multi\_surface}) & 8/50 (16.0\%) & 8/50 (16.0\%) & 8/50 (16.0\%) & 0/42 (0.0\%) \\
Middleware (\texttt{fresh\_middleware}) & 6/50 (12.0\%) & 6/50 (12.0\%) & 6/50 (12.0\%) & 0/44 (0.0\%) \\
Listeners (\texttt{fresh\_listeners}) & 0/50 (0.0\%) & 32/50 (64.0\%) & 0/50 (0.0\%) & 32/50 (64.0\%) \\
\midrule
\textbf{Total Fresh Cohort} & \textbf{105/300 (35.0\%)} & \textbf{152/300 (50.7\%)} & \textbf{105/300 (35.0\%)} & \textbf{47/195 (24.1\%)} \\
\bottomrule
\end{tabular}
}
\end{table}

\section{Statistical Analysis}
\label{app:statistics}

\subsection{Executed Admission Rule and Post-Hoc Rescoring}
\label{app:admission_audit}
The primary 197-task protocol independently evaluates 20 hidden IID and 20 hidden OOD states. For each split,
\[
\operatorname{LCB}_{\mathrm{Wald}}(k,n)=\max\!\left(0,\hat p-1.96\sqrt{\frac{\hat p(1-\hat p)}{n}}\right),\qquad \hat p=k/n.
\]
The critical value $z=1.96$ is the two-sided 95\% normal critical value. Admission requires both split endpoints to be at least $\tau_R=0.85$, in addition to the development and language gates. At $n=20$, 20/20 gives 1.0000 and 19/20 gives approximately 0.85448 (both pass); 18/20 gives approximately 0.76852 (fails). Thus the hidden rule is $k_{\mathrm{iid}}\ge19$ and $k_{\mathrm{ood}}\ge19$.

The earlier Wilson description was a reporting error. A post-hoc robustness check pooled the two splits ($N=40$) and applied the lower endpoint of the two-sided 95\% Wilson interval at the same 0.85 threshold. All recorded candidate evaluations retained identical admission decisions, including 180/197 rescues under $D_0L_1$, with zero stratum migrations. The rules are not mathematically identical: $(19,19)$ passes split Wald but fails pooled Wilson; that combination did not occur in the recorded candidate evaluations. Pooled Wilson was solely post-hoc rescoring of the primary protocol. The separate 100-state fresh-holdout Wilson analyses in Appendix~\ref{app:fresh} retain their stated scope.

\subsection{Paired Contrasts}

For the three protocol-locked contrasts, we use paired two-sided exact McNemar tests. Let $b$ and $c$ denote discordant pair counts. The exact test uses $X\sim\operatorname{Binomial}(b+c,0.5)$ and $p = 2\Pr\left[X\leq\min(b,c)\right]$, capped at one. Holm--Bonferroni correction is applied jointly across $C_1, C_2, C_3$. Paired risk-difference confidence intervals are estimated with 100,000 paired bootstrap resamples.

The final confirmatory results are:
\begin{itemize}
    \item $C_1$: $\Delta = +79.17$ pp, 95\% CI $[+66.67, +89.58]$, $p_{\mathrm{raw}}=7.28\times 10^{-12}$, $p_{\mathrm{Holm}}=1.46\times 10^{-11}$.
    \item $C_2$: $\Delta = +99.30$ pp, 95\% CI $[+97.90, +100.00]$, $p_{\mathrm{raw}}=3.59\times 10^{-43}$, $p_{\mathrm{Holm}}=1.08\times 10^{-42}$.
    \item $C_3$: $\Delta = -6.29$ pp, 95\% CI $[-11.19, -2.10]$, $p_{\mathrm{raw}}=p_{\mathrm{Holm}}=0.01171875$.
\end{itemize}

\section{Robustness Analysis and Execution Integrity}
\label{app:integrity}

\subsection{Family-Level Leave-One-Out Sensitivity}
Leave-one-family-out estimates remain strongly positive for $C_1$ and $C_2$. For $C_3$, the estimated risk difference remains negative across all leave-one-family-out partitions. The exploratory full-cohort factorial interaction retains its negative sign ($I=-25.89$ pp, 95\% CI $[-34.01, -18.27]$).

\subsection{Execution Integrity and Protocol Locking}
The experimental implementation uses fail-closed execution. Model-service failures abort execution, malformed outputs consume an attempt, and recovery primitives outside the assigned language are rejected before execution. Result artifacts, task manifests, protocol configurations, success vectors, and analysis scripts are stored with cryptographic hashes to distinguish frozen result artifacts from superseded runs.
 
\end{document}